\documentclass[10pt,letterpaper]{article}

\usepackage{cogsci}

\cogscifinalcopy 

\usepackage[
    backend=biber,
    style=apa,
    natbib=true,
    doi=false,
    isbn=false,
    url=false,
]{biblatex}
\usepackage{pslatex}
\usepackage{float} 

\usepackage{amsmath}
\DeclareMathOperator*{\argmax}{arg\,max}

\DeclareMathOperator*{\RR}{RR}

\usepackage{amsmath,amssymb}

\usepackage{xcolor}
\usepackage{hyperref}
\usepackage{graphicx}
\usepackage{hyperref}

\title{Toward a normative theory of (self-)management by goal-setting}

\author{{\large \bf Nishad Singhi (nishad.singhi@tuebingen.mpg.de)}  \\ Max Planck Institute for Intelligent Systems, T\"ubingen, 72076, Germany
  \AND {\large \bf Florian Mohnert (mail@florian-mohnert.de)}
   \AND {\large \bf Ben Prystawski (benpry@stanford.edu)}, Department of Psychology, Stanford University, CA, USA
  \AND {\large \bf Falk Lieder (falk.lieder@tuebingen.mpg.de)} \\
 Max Planck Institute for Intelligent Systems, T\"ubingen, 72076, Germany}

\begin{document}

\maketitle

\begin{abstract}

People are often confronted with problems whose complexity exceeds their cognitive capacities. To deal with this complexity, individuals and managers can break complex problems down into a series of subgoals.
Which subgoals are most effective depends on people's cognitive constraints and the cognitive mechanisms of goal pursuit. This creates an untapped opportunity to derive practical recommendations for which subgoals managers and individuals should set from cognitive models of bounded rationality. 
To seize this opportunity, we apply the principle of resource-rationality to formulate a mathematically precise normative theory of (self-)management by goal-setting. We leverage this theory to computationally derive optimal subgoals from a resource-rational model of human goal pursuit. Finally, we show that the resulting subgoals improve the problem-solving performance of bounded agents and human participants.
This constitutes a first step towards grounding prescriptive theories of management and practical recommendations for goal-setting in computational models of the relevant psychological processes and cognitive limitations. 

\end{abstract}

\section{Introduction}
Many complex problems require planning many steps ahead. People often struggle with such problems because their capacity for planning is limited \citep{prystawski2022resource}. To overcome this challenge, individuals, managers, and educators often break complex problems down into a series of subgoals \citep{simon1975functional,catrambone1998subgoal,drucker2012practice}.

In principle, it should be possible to use models of human planning \citep[e.g.,][]{callaway2022rational}, goal-pursuit\citep[e.g.,][]{prystawski2022resource}, and problem-solving \citep[e.g.,]{newell1972human,anderson2013architecture} to predict which subgoals are most beneficial for people. This creates an untapped opportunity to derive practical recommendations for which subgoals managers and individuals should set from cognitive models of bounded rationality \citep{lieder2020resource,lewis2014computational,gershman2015computational}. 
Here, we formulate a mathematically precise normative theory of (self-)management by goal-setting that makes it possible to derive optimal subgoals from computational models of bounded rationality \citep{lieder2020resource}. The basic idea is that goal-setting serves to decompose a complex problem into a series of simpler problems that require less planning. To apply this theory, we combined it with a computational model of human goal-pursuit and an optimization algorithm.

This approach allowed us to improve the problem-solving performance of bounded agents and crowd workers in simulations and an online experiment, respectively. These findings suggest that it might be possible to ground prescriptive theories of managerial and personal goal-setting in computational models of bounded rationality.

The outline of this paper is as follows: We first introduce the relevant theoretical background. We then propose our normative theory of goal-setting for (self-)management. The following two sections evaluate our theory in simulations and an online experiment. We close by discussing our results, their implications, limitations, and directions for future work.

\section{Background}

\subsection{Computational models of bounded rationality}
Human cognition is constrained by bounded cognitive resources and having to solve complex problems in a limited amount of time \citep{lieder2020resource}. To deal with this, the brain uses heuristic strategies that can perform reasonably well with limited resources. The framework of resource rationality \citep{lieder2020resource} formalizes this intuition and seeks to understand human behavior as the optimal use of limited cognitive resources and information. Formally, the resource-rational heuristic that people should use in an environment $E$ is
\begin{equation*}
    h^\star = \argmax_{h \in H_B} \mathbb{E} [\RR(h, E, B)]
\end{equation*}
where $H_B$ is the set of all heuristics that can be implemented by brain $B$ and the resource rationality $\RR$ of a heuristic $h$ implemented by brain $B$ in an environment $E$ is 
\begin{multline*}
    \RR(h, E, B) = \mathbb{E}_{P(\textrm{result} | s_0, h, E, B)}[u(\textrm{result})] \\ - \mathbb{E} [\textrm{cost}(t_h, \rho) | h, s_0, B, E]
\end{multline*}
where $u(\textrm{result})$ is the person's subjective utility of the result obtained by using the heuristic $h$ in situation $s_0$ and $\textrm{cost}(t_h, \rho)$ is the total opportunity cost of investing resources $\rho$ used by heuristic $h$ for time $t_h$. 

Resource-rational models take into account cognitive limitations and perform as well as possible under those constraints. This framework has been successfully applied to various cognitive processes such as planning \citep{callaway2022rational}, goal pursuit \citep{prystawski2022resource}, and decision-making \citep{bhui2021resource}.

\subsection{Simulated Microworlds}
Simulated Microworlds (SMWs) are used to study problem-solving in naturalistic, dynamic environments \citep{funke1993microworlds}. They consist of various variables that interact according to a dynamical system with discrete time steps. The agent can directly manipulate a subset of the variables while only indirectly influencing the values of other variables. For instance, the ultimate aim of the owner of a bakery is to maximize profits. However, she cannot directly control her profit, nor can she influence other factors like demand and sales. Instead, she has to decide how much to spend on advertising, rent, wages, raw materials, etc. in a way that would lead to maximum profit. This makes SMWs a suitable paradigm to study problem-solving in the real world. 

In this study, we adopt the SMW introduced by \citet{mohnert2019testing}, in which participants manage a farm. Their goal is to bring various crops (also referred to as `states') close to specified target values by adjusting how much fertilizer and pesticides (also referred to as `actions') are deployed on the fields (Fig.~\ref{fig:no_subgoal_sc}). In this SMW, participants have complete knowledge of the relationships between all variables, and suboptimality in their behavior can therefore be attributed to their limited cognitive resources. The dynamics of the system are described by the following equation:
\begin{equation*}
\textbf{s}_{t+1} = \textbf{f}(\textbf{s}_t, \textbf{a}_t) = \textbf{A}\textbf{s}_t + \textbf{B}\textbf{a}_t
\end{equation*}
where $\textbf{s}_t \in \mathbb{R}^N$ is the \textit{state} containing values of various crops at time $t$, $\textbf{a}_t \in \mathbb{R}^M$ is the \textit{action} consisting of the amounts of various resources used at time $t$, $\textbf{A} \in \mathbb{R}^{N \times N}$ determines how various crops affect themselves and each other from time $t$ to $t+1$, and $\textbf{B} \in \mathbb{R}^{N \times M}$ determines how resources affect crops.

\subsection{Resource Rational Models of Goal Pursuit}
In simulated micro-worlds, participants often struggle to achieve the task's ultimate goal \citep{prystawski2022resource, funke1993microworlds} because doing so requires planning many steps ahead and taking various factors into account, which can be cognitively demanding. To effectively model how people pursue goals in SMWs, it is important to consider these cognitive limitations. To this end, \citet{prystawski2022resource} developed resource-rational models of goal pursuit, which accommodate limits on people's attention and how many steps they can plan ahead. They found that the model that best explained the problem-solving behavior of the largest proportion of participants in the SMW described above was a hill-climbing model inspired by Newell and Simon's foundational research on human problem-solving \citep{newell1972human}.

\subsubsection{The Hill-Climbing model}
This model assumes that the agent's limited computational resources prevent it from planning more than one step into the future. The agent, therefore, chooses the combination of inputs (actions) that maximally reduces the distance to the goal in the very next round, while also minimizing the cost of those inputs. Any input can be described in terms of the direction of the change in the system's state and the distance between the previous state and the next state. Since the goal of the agent is to minimize its distance from the goal, the agent moves in the direction opposite to the gradient of this distance. Concretely, the action chosen by the agent is:
\begin{equation*}
\textbf{a}_t = - \lambda \cdot \lambda_{opt} \cdot \nabla_{\textbf{a}} || \textbf{f}(\textbf{s}_t, \textbf{a}) - \textbf{g} ||_2
\end{equation*}
where $\lambda_{opt}$ is the optimal step size\footnote{The optimal step size is the step size that minimizes the distance from the goal immediately after a single step in the optimal direction. It is obtained by setting the derivative of the distance from the goal equal to zero while keeping the direction fixed.} in the direction of the negative gradient, $\textbf{g}$ is the goal, $\textbf{f}(\textbf{s}_t, \textbf{a})$ is the state of the agent after taking action $\textbf{a}$ in the state $\textbf{s}_t$. The gradient is evaluated at $\textbf{a} = 0$. $\lambda$ is a free parameter that captures people's tendency to take steps that are systematically smaller or larger than the optimal step size. Stochasticity in people's actions is captured via noise added to the distance and the direction of the agent's actions (see \citet{prystawski2022resource} for details).

\subsection{The resource-rational perspective on goal-setting}
It has been proposed that the function of goals is to reduce the amount of planning that is necessary to reach good decisions \citep{lieder2020resource}. According to a recent refinement of this perspective \citep{correa2020resource,correa2022humans}, an optimal sequence of subgoals should minimize the sum of the costs of the actions the person will take and the mental effort they have to invest into planning to select those actions.

\subsection{Goal-setting in management and self-regulation}
A common approach to management is \textit{management by objectives} \citep{drucker2012practice}. In this approach, the manager's first responsibility is to break down a complex problem into a series of subgoals (\textit{objectives}) that their subordinates can achieve efficiently. The manager assigns their subordinates one subgoal at a time. The employees then work towards the assigned subgoal, and once they accomplish it, the manager assigns them the next subgoal. The purpose of this management practice is to enable teams to achieve challenging long-term goals whose achievement requires considerable amounts of planning and problem-solving. Numerous studies have consistently found that organizational productivity benefits from management by objectives in general \citep{rodgers1991impact} and its goal-setting component in particular \citep[e.g.,][]{mento1987meta}. 

Moreover, people can also use goal-setting to manage themselves, improve their own performance, and help themselves achieve their long-term goals \citep{carver2001self,latham1991self,zimmerman2012goal}.

\section{A normative theory of goal-setting for (self-)management}

The finding that people routinely improve the performance of others (or themselves) through goal-setting raises several interesting, interrelated questions about which subgoals are most effective, what makes them so effective, and how supervisors managing teams (and individuals managing themselves) can generate them. Here, we approach this question from the perspective of rational analysis \citep{anderson2013adaptive} by formalizing the problem people managing others or themselves solve by setting objectives. For the ease of reading and understanding, we describe the theory for the case of managerial goal-setting. But, in principle, the theory also applies when the person being managed is the manager herself.  


Based on the research on goal pursuit summarized in the previous section \citep{prystawski2022resource}, a key problem that managers have to solve is that the path to achieving the organization's ultimate goals may be very long and complex relative to individual employees' capacity and/or propensity for planning. From this perspective, managerial goal-setting serves to reduce the amount of planning that is required for effective goal pursuit \citep[cf. ][]{correa2020resource,correa2022humans}. This, in turn, increases employees' performance on problems that are beyond the cognitive capacity of any single individual. From this perspective, the function of managerial goal-setting is to translate a long-term goal into a sequence of subgoals such that when employees devote their limited cognitive resources to the pursuit of those subgoals they will make more progress towards the long-term goal than they would if they pursued the long-term goal directly. 

To formalize this idea, we model employees' decisions using the recently developed model of boundedly rational goal pursuit introduced above \citep{prystawski2022resource}. Concretely, we model the employee as an agent $\alpha$ that interacts with an environment $E$ by taking action $\textbf{a}_t \in \mathbb{R}^M = (a_t^0, a_t^1, \ldots, a_t^M)$ at time $t$ based on the environment's state $\textbf{s}_t \in \mathbb{R}^N = (s_t^0, s_t^1, \ldots, s_t^N)$. A goal $g$ is characterized by its target values $\textbf{s}_g \in \mathbb{R}^N$, threshold $\delta_g \in \mathbb{R}$, and scale $\gamma_g \in \mathbb{R}^{N \times N}$. The scale captures the possibility that it is more important to bring some state variables closer to their target values than others, and the threshold specifies how close the agent needs to be to the target values to successfully reach the goal. The agent is considered to have achieved the goal if 
$
\sqrt{(\textbf{s} - \textbf{s}_g)^\intercal \gamma_g (\textbf{s} - \textbf{s}_g)} \leq \delta_g, 
$
where $\gamma_g$ is a diagonal matrix with its $i$\textsuperscript{th} value being equal to the scale of the $i^{th}$ state variable. Because values of the scale parameter can be hard to interpret for humans, we converted them to tolerance values for human participants in our experiment. The tolerance $\theta_i$ for the $i^{th}$ state variable is computed as
$\theta_i = \frac{\delta_g}{\sqrt{N \cdot (\gamma_g)_i}} \refstepcounter{equation}(\theequation) \label{eq:scale2tolerance}$, 
where $(\gamma_g)_i$ is the $i^{th}$ diagonal element of $\gamma_g$.

Subgoals are defined similarly, but for a subset of state variables. Concretely, a subgoal $\epsilon$ can be defined for a set of state variables $D_{\epsilon} = (i_1, i_2, ..., i_d)$ with target values $\textbf{s}_{\epsilon}$, scale $\gamma_{\epsilon}$, and threshold $\delta_{\epsilon}$. The condition for successfully achieving a subgoal is
$
\sqrt{(\textbf{s}_r - \textbf{s}_{\epsilon})^\intercal \gamma_{\epsilon} (\textbf{s}_r - \textbf{s}_{\epsilon})} \leq \delta_{\epsilon}, 
$
where $\textbf{s}_r$ is the \textit{reduced state} which is obtained by considering the state variables of $\textbf{s}$ that are included in $D_{\epsilon}$, i.e., $\textbf{s}_r = (s^i | i \in D_{\epsilon})$. Scales for subgoals can be converted to tolerances as done in Eq.~\ref{eq:scale2tolerance} by replacing $N$ with the size of $D_\epsilon$, $\delta_g$ with $\delta_\epsilon$, and $\gamma_g$ with $\gamma_\epsilon$.

The agent starts at $t = 0$ from state $\textbf{s}_0$ and tries to achieve a series of subgoals $\epsilon_1, \epsilon_2, \ldots, \epsilon_k$ and subsequently the final goal $g$. Let $G = (\epsilon_1, \epsilon_2, \ldots, \epsilon_k, g)$. The trajectory $\tau$ of the agent is $\tau = (\textbf{s}_0, \textbf{a}_0, \textbf{s}_1, \textbf{a}_1, \ldots, \textbf{s}_{T-1}, \textbf{a}_{T-1}, \textbf{s}_T)$, where $T$ is the duration of the trial. The expected performance of the agent $\alpha$ is the expected value of the quality $\phi(\tau, g, E)$ of potential trajectories $\tau$ across all trajectories that might occur, that is 
\begin{equation}
\label{eq:performance}
\phi_{\textbf{s}_0}^{G}(\alpha) = \mathbb{E}_{P(\tau | \textbf{s}_0, G, \alpha, E)} \left[\phi(\tau, g, E)\right].
\end{equation}

In principle, more subgoals would lead to better performance. But, in practice, the desired number of subgoals ($k$) is limited by the manager's time, the frequency of the manager's communication with the employees, and the employee's need for autonomy \citep{ryan2006self}.

With all of these definitions in place, we can now define the optimal solution to the problem of managerial goal-setting as selecting the sequence of subgoals $\epsilon_1^\star, \epsilon_2^\star, \ldots, \epsilon_k^\star$ that maximizes the employee's expected performance given uncertain knowledge about its capacities
($P(\alpha)$) as
\begin{equation}
\label{eq:optimalsubgoals}
\epsilon_1^\star, \epsilon_2^\star, \ldots, \epsilon_k^\star = \argmax_{\epsilon_1, \epsilon_2, \ldots, \epsilon_k} \mathbb{E}_{P(\alpha)}[\phi_{\textbf{s}_0}^{G}(\alpha)].
\end{equation}

If the normative theory of managerial goal-setting formalized in Equation~\ref{eq:optimalsubgoals} is potentially useful, then subgoals derived from this theory should improve worker's performance. In the following two sections, we test this prediction with simulated and real workers, respectively.

\section{Improving the performance of bounded agents}

We test our normative theory of managerial goal-setting in the SMW described in the following paragraph. In this section, we apply the theory to compute one optimal subgoal for a resource-rational model of goal-pursuit and check if it improves the model's problem-solving performance.  

\paragraph{The management problem: maximizing the productivity of a farm}
The simulated micro-world used in the present study simulates the problem of managing a farm (see Fig.~\ref{fig:no_subgoal_sc}). Starting from the state $s_0$, the goal of the agents is to bring the values of certain crops (corresponding to state, $\textbf{s}_t$; shown on the right in Fig.~\ref{fig:no_subgoal_sc}) close to the target values specified by the manager. Agents can do so over the course of $T = 20$ steps by using various costly resources (corresponding to action, $\textbf{a}_t$; shown on the left in Fig.~\ref{fig:no_subgoal_sc}). Further, agents have complete information about how crops are influenced by resources and each other, as shown by the weighted edges in Fig.~\ref{fig:no_subgoal_sc}. In particular, we studied the setting where agents start from $\textbf{s}_0 = [80, 20, 90, 10, 70]^\intercal$ and pursue the goal with $\textbf{s}_g = [0, 0, 0, 0, 0]^\intercal, \gamma_g = \textbf{I}_5$ (i.e., identity matrix of size 5), $\delta_g = 50$, and $\delta_{\epsilon} = 1$.

The performance of the agent is higher the larger the number of time steps for which it achieves the goal. Since resources are costly, using more resources leads to lower performance. We formalize this using the \textit{goal-achievement score} (GAS). If the agent achieves the goal for $x$ out of $T$ time steps, and $y = \sum_t ||\textbf{a}_t||_1$, then GAS is defined as
\begin{equation}
\textrm{GAS}(\tau, g, E) = \textrm{max}(0, w_1 + w_2 \cdot x - w_3 \cdot y) \label{eq:GAS}
\end{equation}
where $w_1$, $w_2$, and $w_3$ capture the starting endowment, the reward for achieving the goal, and the cost of resources, respectively. Here, we used $w_1 = 0.2, w_2 = 0.3, w_3 = 0.005$ to capture that achieving the goal is most important. The aim of the agent is to maximize its GAS, while the aim of the manager is to provide a sequence of subgoals to the agent such that pursuing them helps it maximize its GAS.

An important feature of this environment is that the state variable Crowding has an edge with weight = +1.5 starting and ending in it, which creates a positive feedback loop. Without intervention, this feedback loop would cause the value of Crowding to increase exponentially over time. To prevent this, the agent has to bring the value of Crowding close to 0. We therefore predicted that a good subgoal should include Crowding = 0.

\paragraph{Computing an (approximately) optimal subgoal for boundedly rational employees}
To approximate the optimal goal defined in Equation~\ref{eq:optimalsubgoals}, we approximated the expectation in Eq.~\ref{eq:performance} by running $\eta$ noisy simulations of the given agent. We approximated the expectation in Equation~\ref{eq:optimalsubgoals} by averaging the performances ($\phi_{\textbf{s}_0}^{G}(\alpha)$ in Eq.~\ref{eq:optimalsubgoals}) over a population ($\omega$) of hill-climbing agents with different values of step size $\lambda$ that covered the behavior of the largest proportion of participants in \citet{prystawski2022resource}.  To derive $\omega$, we first selected participants from \citet{prystawski2022resource} which were best explained using the hill-climbing model, arranged their step sizes (the only free parameter) in ascending order, and selected 30 equally spaced step sizes to cover the entire range of participants. For each agent, we measured the quality of its trajectories by the goal achievement score defined in Eq.~\ref{eq:GAS}.

To further simplify the computational problem, we considered only one 2-dimensional subgoal (i.e., $k = 1$) and used $\eta = 1$. The distance and angular noise in the actions of the hill-climbing agent were drawn from an exponential distribution (with intensity parameter, $\nu = 0.1$) and a von Mises distribution (centered at $0^\circ$ with concentration parameter $\kappa = 40$), respectively. 

We then approximated the optimal subgoal defined in Eq.~\ref{eq:optimalsubgoals}, by maximizing the simulated performance using Cross-Entropy (CE) Optimization \citep{de2005tutorial}. We ran a separate Cross-Entropy procedure for every possible pair of state variables to optimize for $\textbf{s}_{\epsilon}$ and $\gamma_{\epsilon}$, and then chose the state variables with the highest performance. We ran the CE procedure for 10 iterations with 1000 candidate subgoals in every iteration and selected the top 20\% subgoals in every iteration to refine the distribution of potential subgoals.

\paragraph{Results}
The best subgoal according to our subgoal discovery procedure was Crowding = 0 and SpaceWorms = 4 with the $2 \times 2$ diagonal matrix having elements 0.121 and 0.012 respectively as the scale parameter $\gamma_\epsilon$. This scale parameter translates to tolerance values of ±2 and ±6 for the two subgoal variables, respectively. This is consistent with our prediction that a good subgoal should include Crowding = 0. Simulations with $\eta=100$ showed that the goal-achievement scores of agents are higher in the presence of subgoals vs. without them according to a Mann-Whitney U-test (0.696 vs. 0.069, $U = 5.9 \times 10^6, p < 0.001$). This highlights the efficacy of the subgoal and our procedure. The standard deviations of the two target values in $\textbf{s}_\epsilon$ across 5 runs of the subgoal discovery procedure were 0.71 and 1.41, respectively, which shows that the subgoals discovered by our method are reproducible. 

\section{Improving the performance of crowdworkers}
We performed a pre-registered experiment to test if the subgoals generated by our method help people in problem-solving. To do so, we tested if providing the subgoal computed in the previous section improves people's performance in the simulated micro-world described above (Figure~\ref{fig:no_subgoal_sc}). Participants in the subgoal condition were asked to pursue the subgoal computed by our method before pursuing the final goal (Figure~\ref{fig:subgoal_sc}), whereas participants in the control condition were directly asked to pursue the final goal (Figure~\ref{fig:no_subgoal_sc}). The pre-registration is available at \href{https://aspredicted.org/5W2_GTV}{\texttt{https://aspredicted.org/5W2\_GTV}.}

\subsection{Methods}
\paragraph{Participants}
We recruited 441 crowd workers from the online study platform \textit{Positly}, out of which 234 identified as male, 190 identified as female, and 17 chose not to disclose their gender. The minimum and maximum ages reported were 21 and 76, respectively, with the average age being 40.28 ($SD = 12.05$). Participants spent an average of 39.7 minutes in the experiment. They earned a base pay of \$3 for completing the training. After this, they participated in a practice trial consisting of six rounds ($T=6$). Participants who achieved the specified goal in this trial were paid \$0.15 and were invited to participate in the main experiment. 302 people participated in the main experiment and received a performance-based bonus, with the average value of the bonus being \$1.28.

\paragraph{Procedure}
At the beginning of the experiment, participants were shown three instruction videos and were given a chance to participate in three practice trials. Then, they had to take a quiz testing their attentiveness and understanding of SMWs. Following this, they were invited to participate in another practice trial of six rounds ($T=6$). Participants who achieved the specified goal in this trial were invited to take part in the main experiment. We only analyzed data from the main experiment, which was completed by 302 participants.

\begin{figure}[h!]
  \begin{center}
  \includegraphics[width=8.5cm]{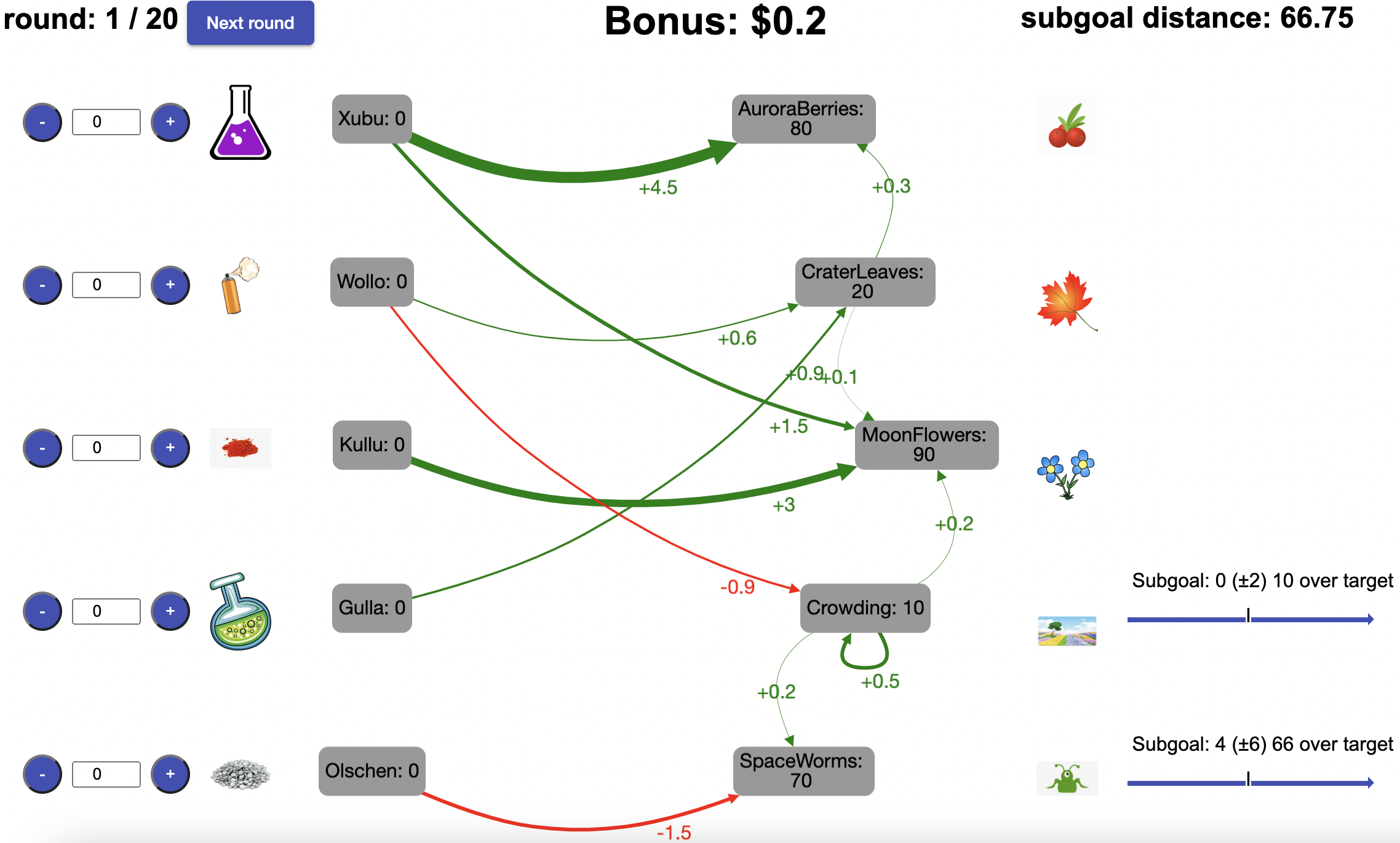}
  \end{center}
  \caption{Screenshot of the Simulated Microworld showing the subgoal discovered by our method. Note that crops not in the subgoal do not have corresponding target and tolerance values because the subgoal has not been achieved yet.}
\label{fig:subgoal_sc}
\end{figure}

We randomly assigned each participant to one of two conditions: the subgoal condition ($n=150$) and the no subgoal condition ($n=152$). Before starting the main experiment, participants in the subgoal condition were informed that they would receive a subgoal that would help them achieve the final goal. Additionally, they were instructed that subgoals can be defined for a subset of crops, and that they should bring the values of these subgoal measures within the specified tolerances of the target values. During the experiment, only the subgoal crops had target values and tolerances next to them (Fig. \ref{fig:subgoal_sc}). Participants were given a message upon successfully achieving the subgoal, following which the subgoal disappeared and the final goal was displayed. 

In the main experiment, which consisted of one trial of 20 rounds (i.e., $T = 20$), participants earned a bonus payment that was equal to their goal-achievement score in USD. Participants in both conditions were informed that they would earn a higher bonus by bringing all crops within their target ranges while using as few resources as possible. They were also informed that negative values of resources have the same cost as positive values. The current value of the bonus was displayed on the screen throughout the trial. In addition, the total distance from the subgoal/final goal was also displayed on the screen.

\begin{figure}[h!]
  \begin{center}
  \includegraphics[width=8.5cm]{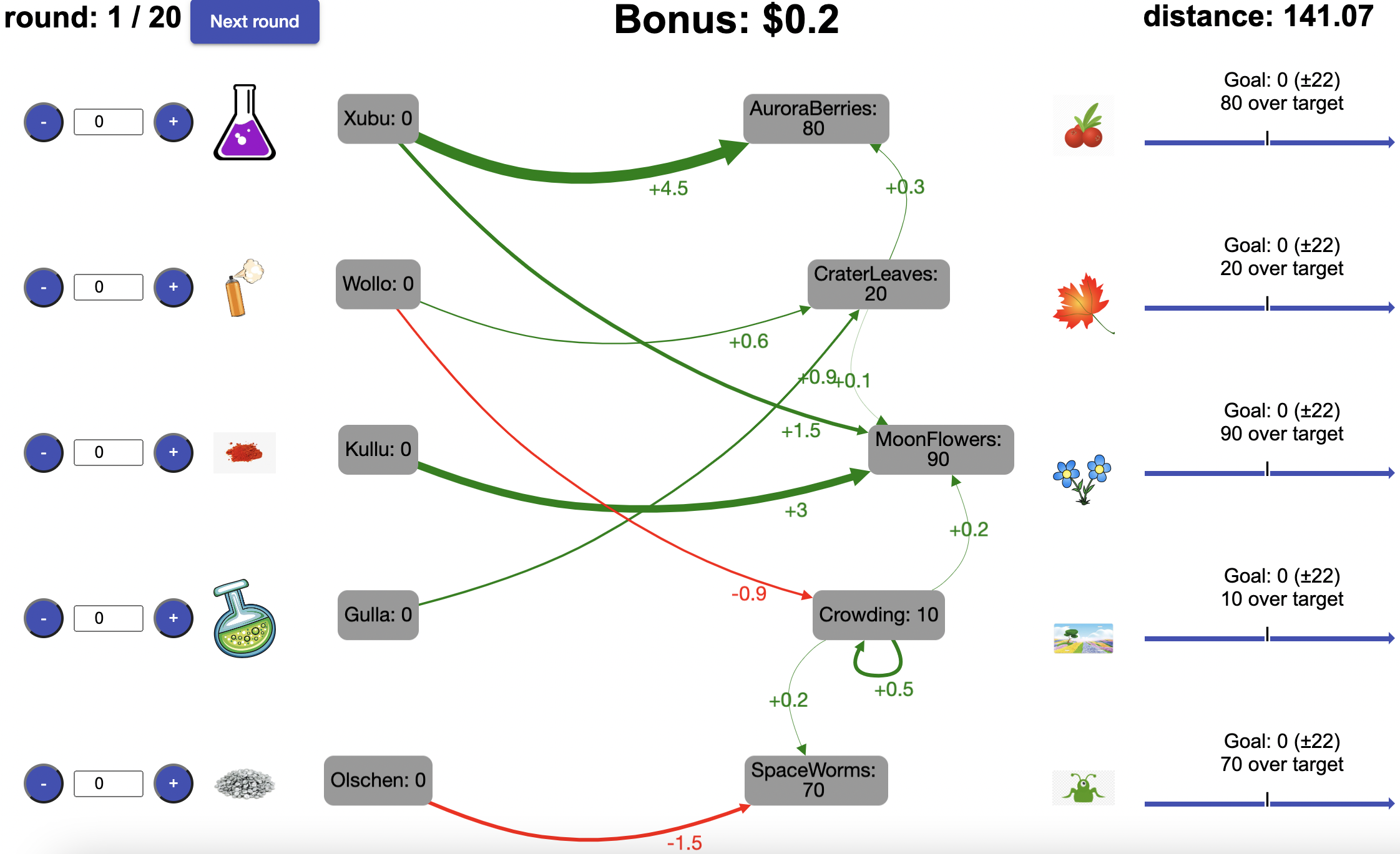}
  \end{center}
  \caption{Screenshot of the Simulated Microworld shown to participants in the experiment.}
\label{fig:no_subgoal_sc}
\end{figure}

\paragraph{Materials}
The experiment involved managing a farm on an alien planet (Fig.~\ref{fig:no_subgoal_sc}) with the goal of bringing the values of certain crops within the specified tolerances of their target values. The target value, tolerance, and current value of each farming measure were shown alongside it on the screen. Participants could influence the values of crops using various costly resources. They could select the amount (positive or negative) for each resource by either typing in the desired value in the corresponding box or by using the up/down arrow keys. The causal relationships between variables were shown via weighted edges. To reduce cluttering, self-connections were only shown when a variable amplified its value over time (Crowding in Fig. \ref{fig:subgoal_sc}).

The starting position and final goal in both conditions were equal to those used previously to compute optimal subgoals. All participants in the subgoal condition were given the subgoal computed by our automatic method: Crowding = 0±2 and SpaceWorms = 4±6. For the practice trials, we employed an SMW which was different from the one used in the main experiment but followed the same rules.

\subsection{Results}
The average goal-achievement scores were 1.33 and 1.23 in the subgoal and no subgoal conditions, respectively. This difference was not statistically significant according to a Mann-Whitney U-test ($U=10595$, $p = 0.11$). However, the proportion of participants with a positive score was significantly higher in the subgoal condition than in the control condition according to a two-proportions z-test (43.33\% vs. 31.6\%, $z= -2.11$, $p=0.035$). 

The amount of resources used by a participant was computed as $\sum_t ||\textbf{a}_t||_1$. We found that participants in the subgoal condition used substantially fewer resources than participants in the control condition (494.45 vs. 859.5, $U= 13092$, $p = 0.012$). 

Additionally, we also computed participants' distance score as
$
\textrm{DS} = \sqrt{||\textbf{s}_T - \textbf{s}_g||_2^2 + c \cdot \sum_{t=0}^{T-1} ||\textbf{a}_t||_2^2}. 
$
The distance score captures how close an agent gets to the final goal at the end of the trial while penalizing it for using more resources. Smaller values of DS are better. Following \citet{prystawski2022resource}, we used $c = 0.01$. In the subgoal condition, the median value of the distance score was numerically lower than in the control condition (492 vs. 988). However, because of the high variance, this difference was not statistically significant according to a one-sided Mann-Whitney U-test ($U= 12466$, $p = 0.08$).
 Taken together with the significant reduction in the amount of resources used, this result indicates that subgoals can help people achieve their goals more efficiently.


In summary, the subgoal condition performed numerically better than the control condition on all outcome measures, was significantly more likely to achieve a positive goal-achievement score, and used the farm's resources significantly more efficiently. These findings are inconclusive, but broadly consistent with the interpretation that the automatically derived subgoal had a small positive effect on people's performance in the problem-solving task.

\section{Discussion}

Goal-setting is commonly used to improve people's ability to solve complex problems \citep{drucker2012practice,catrambone1998subgoal,locke2002building}. One of the reasons why goal-setting is effective is that it reduces the amount of planning that is necessary for goal achievement. It should therefore be possible to leverage resource-rational models of planning and goal-pursuit \citep[e.g.,][]{callaway2022rational,prystawski2022resource} to improve the theory and practice of improving performance through goal-setting \citep{locke2002building}.

To explore this approach, we have proposed a normative theory for (self-)management by goal-setting. We have applied this theory to computationally derive subgoals from a resource-rational model of goal pursuit.
Our proof-of-concept simulations and experiment suggest that it might be possible to derive helpful goal suggestions from resource-rational models of goal pursuit. This illustrates that it is, at least in principle, possible to ground recommendations for goal-setting in the theory of resource-rationality. 
Our work could therefore be considered a first step towards establishing empirically supported computational models of bounded rationality as a micro-foundation for prescriptive theories of (self-)management.

The main limitation of the present work is that empirical evidence for our method's ability to improve human problem-solving was mixed. To explain the mixed results, it is worth noting that the self-amplifying dynamics of the simulated micro-world we used in this experiment made participants' scores extremely variable. This variability, in turn, reduced the power of our statistical tests. As a consequence, even large numerical difference were not always statistically significant. Concretely, the high variability of the scores resulted from the presence of a positive feedback loop that caused Crowding to increase exponentially over time. Once Crowding exceeded a certain value, the exponential growth became unstoppable and participants could no longer control the system. This made the task very challenging for participants, even in the presence of the optimal subgoal. 

Future work should investigate why our participants benefited less from the provided subgoal than our simulations had predicted, and improve the model (and the resulting subgoal) accordingly. 
Follow-up experiments should also investigate the moderating role of individual differences in cognitive ability and motivation.
Moreover, our assessment of the method was limited to a single problem in just one simulated micro-world. Future experiments should assess the generalizability of our findings to other problems in other environments. Such studies could jointly identify under which conditions the subgoals recommended by our method are most beneficial and who benefits the most. 

The work presented in this article builds on, extends, and applies the resource-rational perspective on goals and goal-setting \citep{lieder2020resource}. While previous work explored this idea in planning tasks with discrete states \citep{correa2020resource,correa2022humans}, we have applied the resource-rational perspective on goal-setting to complex problem-solving in dynamic environments with continuous states and inputs. Moreover, while previous work made the unrealistic assumption that the goal is always achieved, our normative theory of goal-setting takes into account that goal-achievement is the exception rather than the norm and that maintaining the desired state is also an important part of the problem. Another innovation of our theory is that it draws on an evidence-based process model of human goal-pursuit, instead of assuming that people use search algorithms that were developed for computers. 

Moreover, while \citet{correa2020resource} and \citet{correa2022humans} studied how people do and machines should decompose tasks into subtasks, our goal was to formulate a normative theory of (self-)management. 
As far as we know, previous research on improving managerial goal-setting did not explicitly engage with mechanistic models of goal pursuit and bounded rationality and did not develop computational methods for computing optimal subgoals.

The work we have presented in this short article can be extended in several directions. One direction is to improve the current method for computing optimal subgoals. Possible improvements include generating a series of multiple subgoals, improving the accuracy and/or speed of the optimization algorithm, considering higher-dimensional subgoals, incorporating the mental effort of planning and goal-pursuit into the objective function \citep[cf.][]{correa2020resource,correa2022humans}.
Further, future work can also use our normative theory of goal-setting as the starting point for resource-rational analyses of how people set goals for others and themselves, respectively. In addition, our work can be extended to make the subgoals more adaptive by learning from the behavior of the employee and adjusting the model of goal pursuit and the subsequent subgoals accordingly.

Finally, we hope that in the long run, the research begun in this project will improve the theory and practice of individual and organizational goal-setting. As a step in this direction, future experiments should compare the effectiveness of the subgoals generated by our method against the effectiveness of subgoals derived from previously proposed heuristics and subgoals chosen by participants. Another step in this direction could be to use an improved version of our theory to generate optimal subgoals for a wide range of complex problems, and then characterize what features useful subgoals have in common.

Although research on grounding prescriptive theories in computational models of bounded rationality is still in its infancy, it is at least beginning to suggest that this is possible to leverage the rigorous methods of computational cognitive science to generate practical knowledge and useful technologies for helping people and organizations become more effective \citep{Lieder2022Interdisciplinary,LiederPrentice_LIS}. We hope that future work in this direction will establish a solid cognitive science foundation for helping people, teams, and organizations set better goals.

\printbibliography

\end{document}